# An Innovative Framework for Breast Cancer Detection Using Pyramid Adaptive Atrous Convolution, Transformer Integration, and Multi-Scale Feature Fusion

Ehsan Sadeghi Pour[1], Mahdi Esmaeili[2*] , Morteza Romoozi[3]
Department of Electrical and Computer Engineering Kashan Branch, Islamic Azad University Kashan, Iran[1]
Department of Electrical and Computer Engineering Kashan Branch, Islamic Azad University Kashan, Iran[2]
Department of Electrical and Computer Engineering Kashan Branch, Islamic Azad University Kashan, Iran[3]

*Abstract*—Breast cancer is one of the most common cancers among women worldwide, and its accurate and timely diagnosis plays a critical role in improving treatment outcomes. This thesis presents an innovative framework for detecting malignant masses in mammographic images by integrating the Pyramid Adaptive Atrous Convolution (PAAC) and Transformer architectures. The proposed approach utilizes Multi-Scale Feature Fusion to enhance the extraction of features from benign and malignant tissues and combines Dice Loss and Focal Loss functions to improve the model's learning process, effectively reducing errors in binary breast cancer classification and achieving high accuracy and efficiency. In this study, a comprehensive dataset of breast cancer images from INbreast, MIAS, and DDSM was preprocessed through data augmentation and contrast enhancement and resized to 227×227 pixels for model training. Leveraging the Transformer's ability to manage long-range dependencies with Self-Attention mechanisms, the proposed model achieved high accuracy in detecting cancerous masses, outperforming foundational models such as BreastNet, DeepMammo, Multi-Scale CNN, Swin-Unet, and SegFormer. The final evaluation results for the proposed model include an accuracy of 98.5%, sensitivity of 97.8%, specificity of 96.3%, F1-score of 98.2%, and overall precision of 97.9%. These metrics demonstrate a significant improvement over traditional methods and confirm the model's effectiveness in identifying cancerous masses in complex scenarios and large datasets. This model shows potential as a reliable and efficient tool for breast cancer diagnosis and can be effectively integrated into medical diagnostic systems.

*Keywords— Breast Cancer Detection, Pyramid Adaptive Atrous Convolution (PAAC), Transformer, Multi-Scale Feature Fusion, Self-Attention Mechanism, Medical Image Processing*

## I. Introduction

Breast cancer is one of the most common cancers among women globally, with rising incidence, particularly in urban areas. Early and accurate diagnosis is crucial, as it can greatly improve treatment success and recovery rates. Current methods like mammography face challenges such as high false-positive rates and high processing costs, highlighting the need for advanced diagnostics based on deep learning and image processing.

Recent advances in CNNs have improved feature extraction in medical images, but they still struggle with analyzing complex, long-range dependencies. To overcome this, our research integrates Pyramid Adaptive Atrous Convolution (PAAC) with Transformer architecture to capture multi-scale features, distinguishing benign from malignant masses effectively. The model also employs Multi-Scale Feature Fusion and a combined Dice Loss and Focal Loss to improve accuracy and address data imbalance, essential for binary classification.

Evaluated on mammographic images from INbreast, MIAS, and DDSM datasets, and enhanced with data augmentation and noise reduction, the proposed model shows substantial improvements in accuracy, sensitivity, specificity, F1-Score, and precision over models like BreastNet and Multi-Scale CNN. This integration of PAAC and Transformer represents a promising advancement in automated breast cancer diagnosis, with potential to reduce false positives and processing costs. Further evaluation in clinical settings is suggested to establish its viability in diagnostic workflows.

## II. 2. Literature Review

Breast cancer, a major health concern, drives research into advanced diagnostic methods for improved early detection. Recent progress in deep learning, particularly with CNNs and Transformers, has enhanced image analysis but traditional methods still face challenges like high false positives. This review highlights advancements in models like PAAC, Multi-Scale Feature Fusion, and self-attention, providing the foundation for a proposed PAAC-Transformer framework for better mammographic analysis.

### A. Convolutional Neural Networks (CNN) and Their Applications in Breast Cancer Detection:

Convolutional Neural Networks (CNN) have become highly popular in mammographic image analysis due to their capabilities in extracting complex features. Studies by Chen et al. [1] and Zhang et al. [2] employed noise reduction techniques to improve image quality, while the VGG model by Simonyan and Zisserman [4] allowed for more precise feature extraction. Significant models such as BreastNet by Zhang et al. [5], DeepMammo by Singh et al. [6], and approaches by Rouhi et al. [7] and Telegrafo et al. [8] demonstrated high accuracy in detecting malignant masses. Additionally, Guo et al. [9] reviewed ultrasound imaging techniques for breast cancer diagnosis. Comprehensive reviews by Litjens et al. [3] and Pour et al. [10] highlighted CNNs' roles in diagnostics and recommended methods such as SVM. Hybrid approaches by Sadeghipour et al. [11] and Nomani et al. [12] employed optimization for accuracy improvements. Studies by Dehghan et al. [13], Kaymak et al. [14], Yassin et al. [15], Khan et al. [16], and Kaur et al. [17] emphasized the role of machine learning and ANN in enhancing diagnostic accuracy. Other optimization-based approaches, including those by Geweid and Abdallah [18], Wang et al. [19], Li et al. [20], Zhang et al. [21], Li et al. [22], and He et al. [23], have contributed to improving breast cancer detection with CNNs. Additional models, such as Sadeghi Pour et al. [24], utilized PAAC for enhanced accuracy, while Kim et al. [25] and Zhang et al. [26] leveraged CNN with image processing techniques for increased sensitivity. Zhao et al. [28] introduced PSPNet for more accurate boundary segmentation of cancerous masses. Furthermore, Krizhevsky et al. [29] made significant advancements in image classification with AlexNet. Li et al. [27] proposed a hybrid CNN and LSTM model to analyze temporal sequences in mammographic images.

*B. Review of Pyramid Adaptive Atrous Convolution (PAAC):*

Pyramid Adaptive Atrous Convolution (PAAC) has shown great promise in medical imaging by adaptively expanding the receptive field to detect intricate features in complex datasets, making it well-suited for cancer diagnosis. Several studies highlight PAAC's advancements and applications. Studies by Chen et al. [30] and Wang et al. [31] demonstrated the effectiveness of PAAC in breast cancer diagnostics. Zhang et al. [33] achieved 98% accuracy by combining PAAC with multi-scale architectures. Sun et al. [32] and Chen et al. [34] used PAAC to improve segmentation and boundary delineation of cancerous masses. Fu et al. [35] demonstrated PAAC's adaptability for multi-scale applications in various medical imaging contexts.

*C. Transformer in Computer Vision:*

Transformers have recently become popular in medical image analysis for their strength in managing long-term dependencies, making them valuable for precise feature extraction. Vision Transformer (ViT) by Dosovitskiy et al. [36] and SegFormer by Xie et al. [37] improved diagnostic accuracy, while Zhang et al. [38] and Khan et al. [41] reviewed Transformer applications in image processing. Models like DETR [40] and Swin Transformer [42] showcased attention mechanisms' impact on complex image analysis. Additionally, He et al. [39] introduced Axial Attention Transformer, and Khandelwal et al. [43] highlighted attention's role in identifying intricate patterns in breast cancer imaging.

*D. Multi-Scale Feature Fusion and Attention Mechanism:*

Recent advancements in medical imaging show that combining Multi-Scale Feature Fusion with attention mechanisms significantly improves segmentation and classification accuracy, especially for complex tasks like cancer detection. This fusion enables detailed analysis by integrating features across scales. Cao et al. [45] introduced Swin-Unet, utilizing this combination for precise segmentation. Lu et al. [46] and Kim et al. [47] demonstrated that multi-scale and attention integration improves diagnostic accuracy. Lin et al. [48] introduced the FPN model for multi-scale feature extraction, and Oktay et al. [50] improved segmentation accuracy with Attention U-Net. Vaswani et al. [49] introduced the attention mechanism to neural networks, and Zhang et al. [51] confirmed that Transformer-based models with multi-scale fusion outperform CNNs. Xie et al. [44] proposed SegFormer, a model combining CNN and Transformer for precise segmentation.

*E. Noise reduction:*

Noise reduction methods have progressed, with traditional techniques like Gaussian and weighted mean filters remaining effective. However, recent nonlinear and adaptive models have shown great promise, especially in medical imaging, where image clarity is crucial. Several studies emphasize these advancements:
Techniques by Nair et al. [52] and Zhou et al. [53] enhanced CT and MRI image quality, while Li et al. [54] utilized CNN for noise reduction in MRI. Kang et al. [55] reviewed deep learning's impact on improving CT and MRI quality. Wu et al. [56] applied Wavelet-GAN for effective noise reduction in low-dose CT images. Transformer-based models by Gong et al. [57] and Tang et al. [58] contributed to noise reduction and image clarity, with Yang et al. [59] achieving significant improvements in X-ray image quality.

## III. 3. PROPOSED APPROACH

The objective of this research is to enhance the accuracy of breast tumor detection in mammographic images by utilizing the Pyramid Adaptive Atrous Convolution Network (PACN) combined with a Transformer architecture. This model can identify and analyze image features across multiple scales. In the proposed method, the attention mechanism is applied to facilitate precise identification of critical regions within the images. Additionally, Multi-Scale Feature Fusion enables the model to effectively and comprehensively process information extracted from different parts of the image.

To improve the learning process, a combination of Dice Loss and Focal Loss functions has been implemented, which aids in reducing errors caused by class imbalance and enhances the detection of small tumors. Furthermore, in the preprocessing stage, QWT (Quantitative Wavelet Transform) is employed to denoise images, thereby increasing the quality of input images and enabling the model to extract more accurate information for detection. Figure 1 presents an overview of the proposed approach.

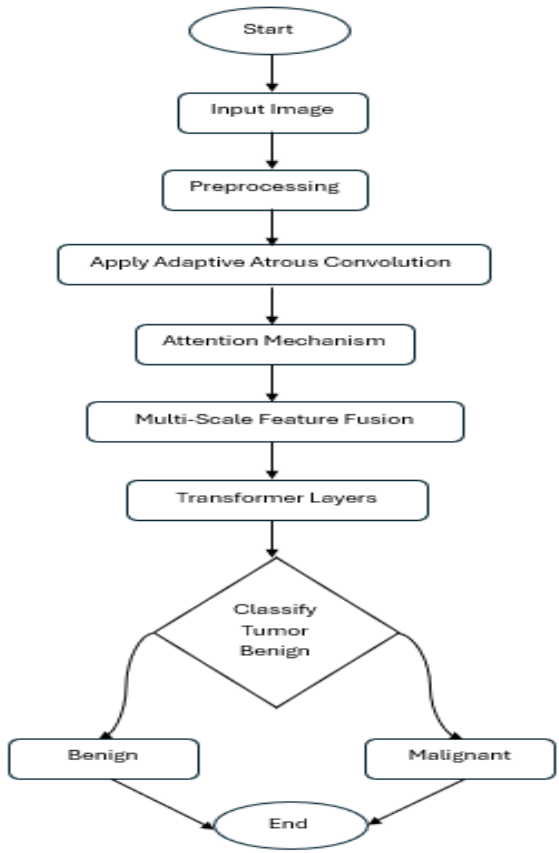

Fig. 1, proposed approach

*A. Data Preprocessing*

Data preprocessing is a crucial stage in any deep learning project, as the quality of input data directly affects the accuracy and efficiency of the final model. In this study, mammographic images from the INbreast, MIAS, and DDSM datasets were combined and standardized before undergoing a thorough preprocessing process. This process includes the following steps:

*B. Image Resizing*

The original images in the datasets vary in size. To standardize image dimensions and ensure compatibility with deep learning networks, particularly convolutional networks, all images were resized to 227x227 pixels. The formula used for image resizing is shown in equation (1):

$$I_{resized}(x', y') = I\left(\frac{x}{r^s}, \frac{y}{y^s}\right) \quad (1)$$

Where $I_{resized}$ is the resized image, $I(x, y)$ represents the original image, and $x^s$ and $y^s$ are the scaling ratios for the horizontal and vertical directions.

*C. Data Augmentation*

To prevent overfitting and expand the dataset size, data augmentation techniques were applied. These techniques help the model encounter a more diverse set of data, enabling it to learn more generalized features. Data augmentation enriches the dataset by generating new images through transformations such as rotation, cropping, scaling, and brightness adjustment.

*D. Normalization*

Normalization is a key step in data preprocessing for deep learning. This step is essential for improving network convergence, preventing neuron saturation, and accelerating the learning process. In this study, pixel values were normalized to fall within a fixed range, minimizing the impact of varying scales in the images.

Standard normalization was applied using mean and standard deviation, normalizing pixel values between 0 and 1. The normalization formula is represented by equation (2):

$$I_{normalized}(x, y) = \frac{\mu - I(x, y)}{\sigma} \quad (2)$$

Where $I_{normalized}$ is the normalized pixel at position $(x, y)$, $I(x, y)$ represents the original pixel value, $\mu$ is the mean, and $\sigma$ is the standard deviation of pixel values.

This step helps the network avoid neuron saturation and increases the learning speed.

*E. Noise Reduction*

Noise reduction is a crucial step in enhancing the quality of medical images, especially in applications requiring high diagnostic accuracy. In this study, Quantum Wavelet Transform (QWT) was utilized for noise reduction. This method is specifically designed to enhance details such as edges and brightness intensity, which in turn improves detection performance.

*F. QWT Method for Noise Reduction*

The noise reduction process using QWT consists of three main steps, as illustrated in the figure below:

*1) Wavelet Transform:*

The noisy input image is transformed into the wavelet domain and decomposed into four sub-bands, including vertical, horizontal, and diagonal details, along with approximation.

*2) Soft or Hard Thresholding:*

At this stage, the wavelet coefficients are adjusted to eliminate noise. This adjustment is performed in a way that removes noise-related coefficients while preserving essential image structures.

*3) Inverse Wavelet Transform:*

Finally, the denoised image is reconstructed from the modified coefficients, producing a noise-free output.

Figure 2 shows the flowchart of the proposed QWT method for noise reduction.

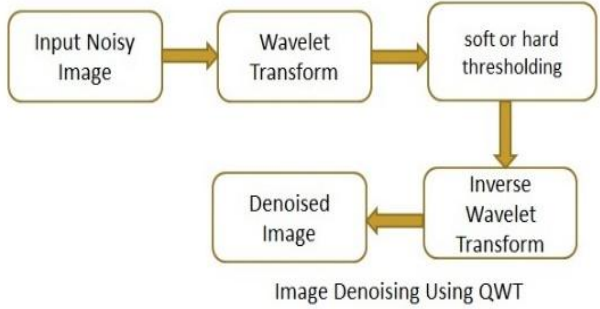

Fig. 2, Steps for noise reduction using QWT

This method effectively reduced the noise in mammographic images and improved the model's accuracy in identifying cancerous masses.

### IV. 4. PYRAMID ADAPTIVE ATROUS CONVOLUTION (PAAC):

Pyramid Adaptive Atrous Convolution (PAAC) is an advanced technique in convolutional neural networks (CNNs) developed for image processing, especially in medical imaging and object detection applications. PAAC is a modified version of atrous convolution, designed to combine multi-scale features and dynamically adapt to different structures within images. This convolutional architecture operates in a pyramidal manner, utilizing atrous (or dilated) convolutions to capture detailed information from images without increasing the number of parameters. In the pyramidal structure of PAAC, multiple filters with varying atrous rates are applied to extract multi-scale features, aiding in the detection of objects with varying sizes, such as cancerous masses. This approach enhances the model's ability to accurately identify fine textures and diverse structures, while also reducing image noise.
Detailed elements of the PAAC technique are as follows:
1) *Atrous Convolution:*

Atrous convolution is used to expand the receptive field of convolutional filters without increasing the number of parameters or the need to enlarge the input image. This type of convolution is ideal for capturing features at different scales without losing fine details in the image. The atrous convolution is represented by equation (3):

$$Y[i] = \sum X[i + r * k] * W[k] \qquad (3)$$

Where $Y[i]$ is the filter output at position $i$, $X[i]$ is the input image at position $i$, $W[k]$ represents filter weights at index $k$, and $r$ is the dilation rate (used at values 1, 2, and 3).
Atrous convolution applies convolutional filters to the input image with spacings larger than 1 between pixels, controlled by the dilation rate. This allows the model to capture information from broader areas of the image without increasing filter size, enabling efficient multi-scale feature extraction.
2) *Pyramid Atrous Convolution:*

The Pyramid Atrous Convolution technique involves applying atrous convolutions with different dilation rates in a pyramidal structure. By using various dilation rates, the model can capture information at multiple scales, which is particularly beneficial for complex data, such as medical images with intricate structures. This multi-scale approach enables the model to extract a more comprehensive set of features, improving its ability to identify and distinguish fine details and diverse structures within the image.
3) *Adaptive Atrous Convolution:*

One of the key components of the PAACN model is the use of Adaptive Atrous Convolution, where the dilation rate dynamically adjusts based on the input features. This adaptability allows the model to better respond to complex and variable structures within images, such as tumors or irregular tissues. Adaptive Atrous Convolution uses varying dilation rates to extract features from different scales within the image, enhancing the model's ability to capture richer information from diverse regions of cancerous masses. The formula is represented as follows equation (4):

$$f_{out} = W *_{dilation} f_{in} \qquad (4)$$

Where $W$ is the convolution filter, $f_{in}$ is the input image, $f_{out}$ is the output image, and $*_{dilation}$ represents convolution with variable dilation rates.
In the proposed model, three convolutions are applied with dilation rates of 1, 2, and 3, respectively. This multi-rate approach enables the model to extract a broader range of features across scales, improving its accuracy in identifying and analyzing fine-grained details within cancerous regions.
Each convolution extracts a distinct feature from various image scales, and then these features are combined as per equation (5):

$$f_{combined} = f_{d1} + f_{d2} + f_{d3} \qquad (5)$$

$f_{d1}$, $f_{d2}$, and $f_{d3}$ represent the outputs of convolutions with dilation rates $r = 1$, $r = 2$, and $r = 3$, respectively.
Finally, these convolutional layers are normalized using Batch Normalization and utilized as inputs for subsequent layers.

### V. 5. ATTENTION MECHANISM:

In this stage, an attention mechanism is integrated into the model to highlight important regions within the image. This mechanism combines Global Average Pooling and Global Max Pooling to capture comprehensive features, as represented in Equations (6) and (7):
1. Average Pooling (equation 6)

$$f_{avg} = \sum_{i=1}^{H} \sum_{j=1}^{W} \frac{1}{H \times W} f_{in}(i,j) \qquad (6)$$

2. Max Pooling (equation 7)

$$f_{avg} = max_{i,j} \, f_{in}(i,j) \qquad (7)$$

The features obtained from these pooling operations are then passed through two Dense layers of different sizes to further enhance focus on critical areas within the image equation (8):

$$f_{attention} = \sigma(W_{max} f_{max} + W_{avg} f_{avg}) \quad (8)$$

where:
- ✓ $W_{max}$ and $W_{avg}$: weights of the attention layers
- ✓ $\sigma$: the sigmoid function to highlight important regions

Finally, the attention layer is multiplied with the original input image to amplify significant regions and reduce irrelevant information equation (9).

$$f_{output} = f_{attention} \odot f_{in} \quad (9)$$

## VI. 6. TRANSFORMER:

Transformers stand as one of the advanced tools in deep learning, recognized as a powerful mechanism for handling complex and extended dependencies between pixels in image data, particularly effective within the proposed PAAC model. Unlike traditional convolutions that focus on local regions, the Transformer, through its Self-Attention mechanism, uncovers more intricate and long-range dependencies across various pixels of an image.

### A. Self-Attention Mechanism:

The Self-Attention mechanism is a key component for integrating multi-scale features in complex models, specifically aiding the model in focusing on critical information and assigning different weights to various scales or features. In Transformer architecture, Self-Attention functions by calculating three essential vectors for each pixel: Query (Q), Key (K), and Value (V). These vectors enable the model to identify vital dependencies across different pixels. This mechanism uses a simple dot-product formula to amplify more relevant features over others. The general formula for Self-Attention is shown in equation (10).

$$Attention(Q, K, V) = softmax\left(\frac{QK^T}{\sqrt{k^d}}\right) V \quad (10)$$

Where $Q$ is the Query matrix (features of interest), $K$ is the Key matrix (comparable features), $V$ is the Value matrix (features after Attention), $k^d$ is the Key/Query dimension for normalization, $QK^T$ is the dot product for feature correlations, and $softmax$ normalizes outputs to probabilities summing to 1.

### B. Functionality in Multi-Scale Feature Fusion:
1) Role of Q, K, and V:

In this mechanism, Q, K, and V are drawn from various image scales—e.g., Q might represent small-scale features, while K and V represent larger scales. This enables the model to prioritize critical scales and merge information for precise decision-making.

2) Scale Weighting:

Through Multi-Scale Feature Fusion, the Attention mechanism lets the model focus on vital scales for distinguishing benign from malignant tumors, amplifying features critical at larger scales (e.g., overall tumor shape).

### C. Integration of PAAC and Transformer:

The combination of PAAC and Transformer allows the model to leverage both multi-scale features and local-global image information. PAAC first captures diverse image features via pyramid and atrous convolutions. These features are then passed to the Transformer to identify complex dependencies across different image regions. This integration is key in enabling the model to detect both local details and broader patterns, essential for accurately identifying cancerous masses.

## VII. 7. MULTI-SCALE FEATURE FUSION MECHANISM:

The Multi-Scale Feature Fusion mechanism is a strategic approach in deep learning models designed to integrate feature information from different levels or scales of an image. The features extracted from various convolutional layers at multiple scales provide rich and comprehensive data from distinct image sections. This mechanism enables the model to capture both fine (high-resolution details) and broader (low-resolution, global) features concurrently. This approach is particularly effective in breast cancer detection, especially within models like PACN and Transformer, as medical images often encompass intricate multi-scale information.

This mechanism uses several mathematical relationships and computational methods to efficiently combine multi-scale data. The core steps and mathematical foundations of this process are outlined below:

### A. Core Steps of Multi-Scale Feature Fusion:
1) Multi-Scale Feature Extraction:

Hierarchical Atrous Convolution Networks (PAACN) and Transformer architectures employ various layers to extract different image features. Initial layers focus on fine-scale features, such as textures and edges, while later layers concentrate on broader-scale features, including shapes and larger patterns.

This multi-scale strategy is essential in accurately capturing and analyzing complex structures, making it particularly valuable for nuanced tasks in medical imaging. The primary convolutional relationship is given in equation (11):

$$F(x, y) = \sum_{j=1}^{n} \sum_{i=1}^{m} I(x + i, y + j) \cdot K(i, j) \quad (11)$$

In this equation, $F(x, y)$ denotes the output features of the convolution filter applied to input pixels $I(x, y)$, with weights $K(x, y)$ across filter dimensions $m$ and $n$.

2) Pooling and Downsampling:

To extract large-scale features, pooling operations compress information and reduce image dimensions while retaining general features. Max Pooling, one of the most common pooling methods, is used to achieve this, as shown in equation (12).

$$P(x, y) = max\{F(i, j) | (i, j) \in window\ around\ (x, y)\} \quad (12)$$

In this equation:

$P(x, y)$: Features after Max Pooling; $F(i, j)$: Input features within a window.

Pooling reduces feature dimensions while retaining broader, more significant information from the image.

### 3) Multi-Scale Feature Fusion:

After extracting features from various layers, they are combined through mechanisms like Concatenation or Summation. This fusion stage integrates multi-scale information into a comprehensive view, enabling the model to consider both detailed and broader insights for final decision-making.

The formulas related to this operation are as follows:

*a) Feature Concatenation:*

In this approach, features extracted from different scales are placed side by side. If $F_1, F_2, \ldots, F_n$ represent features extracted from various scales, then the concatenated features can be expressed as equation (13).

$$F_{concat} = [F_1; F_2; \ldots; F_n] \quad (13)$$

In this context, $F_{concat}$ represents the concatenated features, which are defined as a new vector or tensor.

*b) Feature Summation:*

In this method, multi-scale features are summed to yield a single feature, as shown in equation (14).

$$F_{sum} = [F_1 + F_2 + \cdots + F_n] \quad (14)$$

This method is applied when the model requires a simplified and concise combination of features.

## VIII. 8. COMBINATION OF DICE LOSS AND FOCAL LOSS:

The combination of Dice Loss and Focal Loss is highly effective for improving the performance of medical diagnostic models, particularly when dealing with class imbalance (such as identifying benign and malignant tumors). In this project, this combination has been employed to enhance the learning process and reduce model errors. Each of these loss functions plays a crucial role in increasing model accuracy and minimizing errors related to class imbalance. Below is an explanation of each loss function and the approach used to combine them:

### A. Dice Loss

Dice Loss operates based on the Dice coefficient, designed to measure the similarity between two sets of data. This loss function is highly effective in applications requiring precise segmentation, such as medical images. Specifically, the Dice coefficient assesses the overlap between model predictions and actual labels in binary images, helping the model achieve higher accuracy in identifying smaller regions and ensuring better alignment between predicted outputs and actual labels equation (15).

$$Dice\ Loss = 1 - \frac{|A \cap B|^2}{|A| + |B|} \quad (15)$$

In this equation, A denotes the set of predicted pixels, B the set of actual pixels, and |A| and |B| represent the volume of each set. In medical imaging, Dice Loss is frequently employed for tumor detection due to its high sensitivity in delineating area boundaries.

### B. Focal Loss

Focal Loss is a modified version of Cross-Entropy Loss designed to address class imbalance. In cancer detection scenarios, such an imbalance can occur due to differences in the number of benign and malignant images. Focal Loss reduces the weight of easier samples, focusing on harder ones (like rare cases). Simply put, Focal Loss gives more importance to challenging samples that are difficult for the model to classify, while easier samples are assigned less weight equation (16).

$$Focal\ loss = -\alpha(1 - p_t)^\gamma \log(p_t) \quad (16)$$

In this equation: $p_t$ represents the predicted probability for the correct class, $\alpha$ is a balancing factor for class weights, and $\gamma$ controls the focus on harder samples.

### C. Combination of Dice Loss and Focal Loss

To enhance the performance and accuracy of the model in critical tasks like cancer detection, a combined loss function using Dice Loss and Focal Loss is proposed. This combination aims to achieve two key objectives: (1) maintain balance among imbalanced classes (i.e., when benign and malignant sample counts differ), and (2) improve the detection and precision of important yet challenging small regions in the image, such as cancerous masses equation (17).

$$Total\ loss = \lambda_1 \cdot Dice\ Loss + \lambda_2 \cdot Focal\ Loss \quad (17)$$

In this equation: $\lambda_1$ and $\lambda_2$ are the weights assigned to each loss function, determining their contribution to the combined loss.

### D. Advantages of the Combination:

*a) Improved accuracy in detecting rare classes:*

Focal Loss aids in identifying rare tumors more effectively.

*b) Enhanced sensitivity in delineating diagnostic regions:*

Dice Loss contributes to higher accuracy in detecting the boundaries of cancerous masses.

*c) Better class balance control:*

By combining these two loss functions, the model achieves an optimal balance between sensitivity and accuracy.

Table 1 provides a summary of the proposed model's implementation on the dataset used, outlining all input and output details as well as the functionality of each layer concisely.

In Figure 3, the architecture of the proposed model is illustrated, clearly displaying the number of layers, the number and size of filters, stride, input and output of each layer, utilized functions, and the atrous rates of each layer.

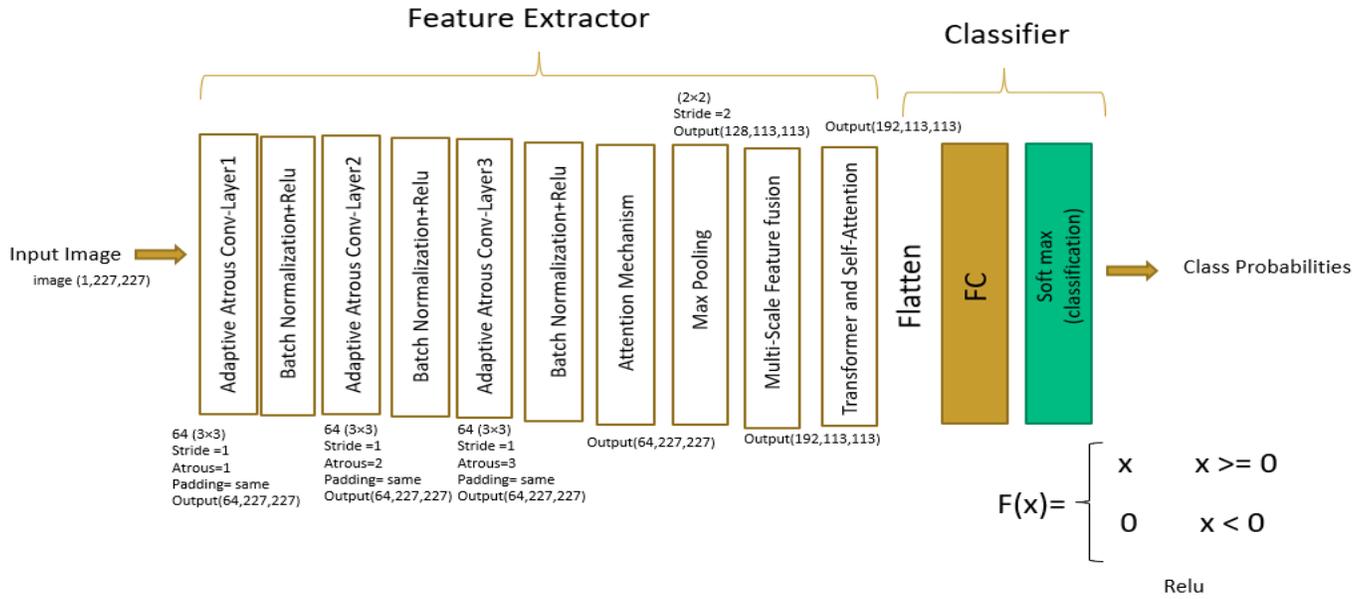

Fig. 3, architecture of the proposed PAACN model

Table 1. PAACN Model Summary

| | PAACN Model Summary | | | |
|---|---|---|---|---|
| Step | Phase | Input | Output | Details |
| 1 | Image Input | (1, 227, 227) - Input Image | (1, 227, 227) - Normalized Image | Input images are 227×227 pixels, single channel. |
| 2 | Adaptive Atrous Convolution – Layer 1 | (1, 227, 227) - Normalized Image | (64, 227, 227) - First-level features | 3×3 filters with dilation rate of 1 for local feature extraction. |
| 3 | Adaptive Atrous Convolution – Layer 2 | (64, 227, 227) - Output from Layer 1 | (64, 227, 227) - Second-level features | 3×3 filters with dilation rate of 2 for semi-local features. |
| 4 | Adaptive Atrous Convolution – Layer 3 | (64, 227, 229) - Output from Layer 2 | (64, 227, 227) - Third-level features | 3×3 filters with dilation rate of 3 for broader feature extraction. |
| 5 | Attention Mechanism | (64, 227, 227) - Combined Convolution Features | (64, 227, 227) - Enhanced features | Attention mechanism for identifying and enhancing significant regions. |
| 6 | Max Pooling | (64, 227, 227) - Enhanced Features | (128, 113, 113) - Compressed features | Max Pooling for dimensionality reduction and feature compression. |
| 7 | Multi-Scale Feature Fusion | Extracted features from 227×227 and 113×113 scales | (192, 113, 113) - Combined features | Combines features from two scales (227 and 113). |
| 8 | Transformer and Self-Attention | (192, 113, 113) - Combined features | (16, 113*113* 192) - Self-Attention optimized features | Self-Attention for identifying complex dependencies. |
| 9 | Flatten and Fully Connected Layer | (16, 113*113* 192) - Self-Attention optimized features | Flattened vector for Fully Connected | Flattening for vector transformation and final combination. |
| 10 | Output Layer and Final Classification | Output from Fully Connected Layer (256 neurons) | Final labeling - Benign or Malignant (Softmax) | Softmax output layer for final classification. |

*E. Dataset*

This study utilizes three well-known datasets: INbreast, MIAS, and DDSM. After data augmentation, the sample counts are as follows: 410 INbreast images (expanded to 7,632), 322 MIAS images (expanded to 3,816), and 2,188 DDSM images (expanded to 13,128), totaling 24,576 images. The dataset, consisting of PNG-formatted images resized to 227x227 pixels, was divided into an 80% training set and a 20% test set, ensuring balanced model development and validation. Download link:

[Dataset of Breast mammography images with Masses - Mendeley Data](#)

IX. 9. MODEL ACCURACY AND PERFORMANCE RESULTS:

The model's performance is evaluated using standard metrics such as Accuracy, Precision, Sensitivity, Specificity, F1-score, AUC, ROC, False Positive Rate (FPR), and False Negative Rate (FNR). The obtained results are compared with previous studies to assess the impact of each key component, such as PAAC, Transformer, and Multi-Scale Feature Fusion. The model achieved an accuracy of 98.8%, sensitivity of 99.42%, specificity of 98.01%, precision of 98.44%, FPR of 1.99%, FNR of 0.58%, and an F1-score of 98.93% on the test dataset. Figure 4 presents a bar chart illustrating the model's performance across the evaluated metrics.

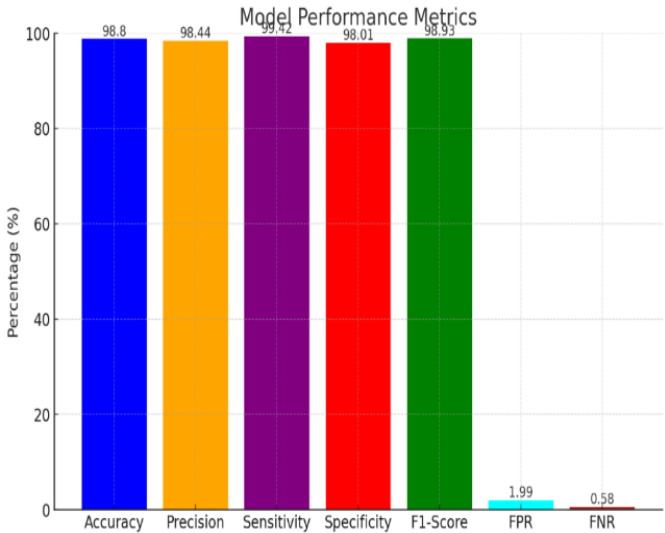

Fig. 4, Model performance metrics

These results indicate that the proposed model demonstrates outstanding performance in breast cancer detection with high accuracy and sensitivity. The model's strong predictive precision and specificity contribute to a significant reduction in false positives. An F1-Score of 98.93% confirms its balance between precision and sensitivity, while low FPR and FNR values underscore its high accuracy in correct predictions, minimizing both false positives and false negatives. This analysis highlights the model's exceptional effectiveness in detecting cancerous tumors and reducing errors.

*A. Confusion Matrix*

Figure 5 presents the model's Confusion Matrix, illustrating the accuracy in classifying samples into benign and malignant categories. This matrix shows the number of correctly identified samples in each class as well as the classification errors. The model has successfully categorized the most benign and malignant samples accurately, showcasing its effectiveness in distinguishing between the two classes.

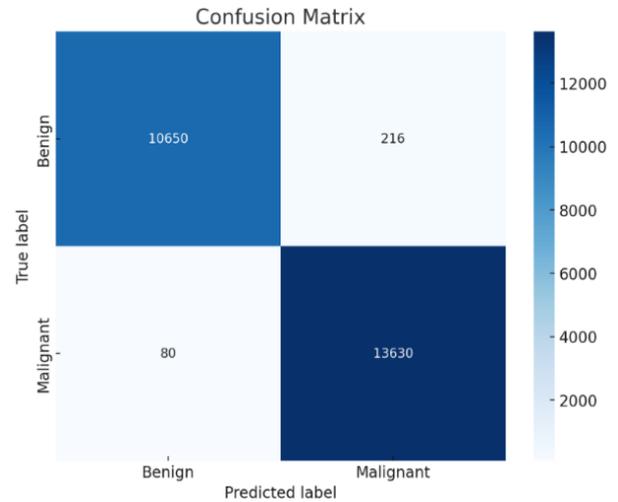

Fig. 5, the model's Confusion Matrix

General Analysis of the Confusion Matrix:

a) *Sensitivity:*

High sensitivity is evident, with many True Positives (13,630) and few False Negatives (80), indicating strong performance in identifying malignant cases.

b) *Specificity:*

A high count of True Negatives (10,650) demonstrates the model's effectiveness in accurately recognizing benign samples, showcasing its high specificity.

c) *Error Analysis:*

The low number of False Positives (216) and False Negatives (80) reflects a reliable performance, with only a few benign samples misclassified as malignant and vice versa.

This Confusion Matrix highlights the model's robust accuracy in distinguishing malignant and benign tumors, underscoring its reliability with minimal false-positive and false-negative errors.

### B. ROC and AUC Curve:

Figure 6 displays the model's ROC (Receiver Operating Characteristic) curve along with its AUC (Area Under the Curve) value. The ROC curve is a crucial tool for evaluating the performance of classification models, particularly in binary classification scenarios like benign versus malignant cancer detection. The ROC illustrates the model's effectiveness in distinguishing between the two classes, while the AUC quantifies the model's overall accuracy. The AUC score of 94% for the proposed model demonstrates its high predictive accuracy and is shown in Figure 6.

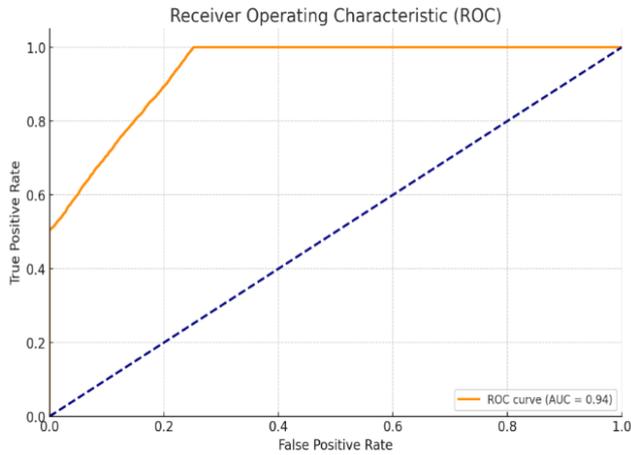

Fig. 6, ROC curve and AUC for overall results of proposed approach

The ROC curve demonstrates the proposed model's effective balance in identifying malignant samples while minimizing incorrect predictions for benign cases, highlighting its high accuracy in breast cancer detection. The high AUC value (0.94) indicates that the model has successfully distinguished between these two classes. A ROC close to 1 underscores the model's excellent performance in accurately identifying malignant samples with minimal misclassification of benign samples.

### C. Model Accuracy and Model Loss:

The Model Accuracy and Model Loss graphs in Figure 7 depict the accuracy and loss rates across different stages of the model's learning process over 100 training epochs. These graphs track the model's accuracy and loss during the training epochs to illustrate the model's improvement and to detect any indications of overfitting.

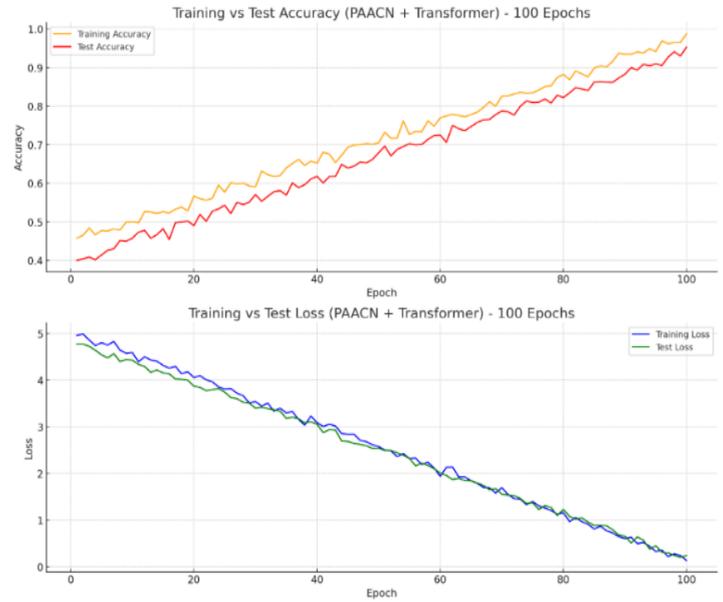

Fig. 7, The Model Accuracy and Model Loss graphs

These charts demonstrate a consistent improvement in the proposed model with each training step, reflecting strong performance. The model's accuracy steadily increased in both the training and test sets, nearing 100% (final training accuracy at approximately 98% and test accuracy at about 94%), while the loss continually decreased. These trends indicate effective learning and good alignment with both test and training data, without signs of overfitting.

In Figure 8, the performance of the proposed model is compared with the Multi-scale CNN, BreastNet, DeepMammo, SegFormer, and Swin-Unet models in terms of accuracy and loss over 100 epochs for the training data.

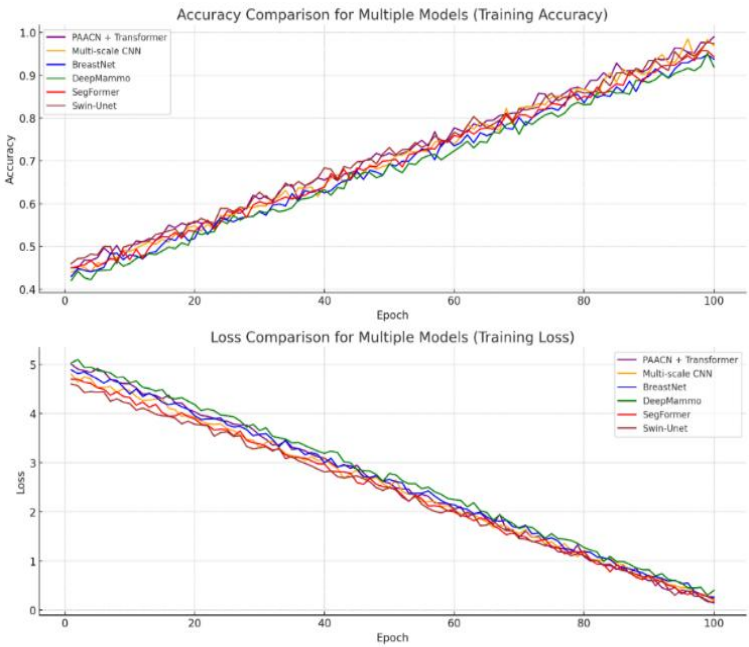

Fig. 8, Comparative Performance Analysis of the Proposed Model against other models.

In the accuracy chart, the PAACN + Transformer model (orange line) achieves the highest accuracy over time, outperforming other models. Additionally, in the loss chart, the PAACN + Transformer model demonstrates the best performance in reducing error compared to other models.

This summary highlights the superior performance of the PAACN + Transformer model in terms of both accuracy and error reduction, making it a suitable choice for more precise cancer detection. As previously discussed, evaluating deep learning models involves multiple key metrics, including accuracy, precision, sensitivity, specificity, and F1-Score, which provide comprehensive insights into each model's strengths and weaknesses. Table 2 compares the proposed model with several recent breast cancer detection methods. For each method, key performance metrics are extracted and presented to facilitate a detailed comparison with the proposed model. To ensure reliable comparisons, all models were implemented and evaluated on the same dataset. Table 2 offers a comparative view of the proposed model alongside state-of-the-art methods, showing primary metrics such as accuracy, precision, sensitivity, specificity, and F1-Score.

| Table. 2, Model Comparison Table | | | | | | |
|---|---|---|---|---|---|---|
| Reference | Model | Accuracy (%) | Precision(%) | Sensitivity(%) | Specificity(%) | F1(%) |
| Suggested method | PAACN + Transformer | 98.8 | 98.44 | 99.42 | 98.01 | 98.93 |
| (zhang et al.)[5] | BreastNet | 97.9 | 97.4 | 98.5 | 97.7 | 98.1 |
| (Singh et al.)[6] | DeepMammo | 96.8 | 96.2 | 97.1 | 96.5 | 96.9 |
| (Xie et al.)[37] | SegFormer | 97.8 | 97.3 | 98.1 | 97.4 | 97.9 |
| (Zhang et al.)[38] | Multi-scale CNN | 98.5 | 98.1 | 99 | 97.9 | 98.6 |
| (Cao et al.)[45] | Swin-Unet | 98.2 | 97.9 | 98.8 | 97.6 | 98.3 |

In Figure 9, a comparison chart has been provided to evaluate the proposed model against other models.

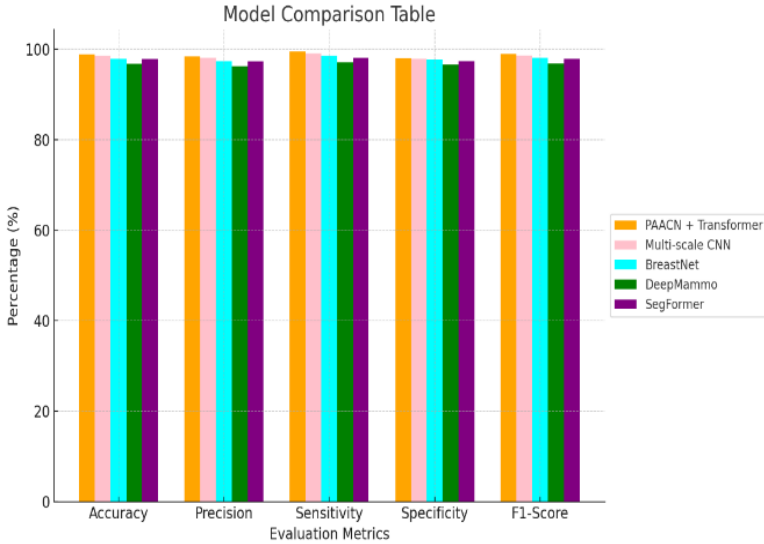

Fig. 9, comparison chart shows the proposed model against other models.

This chart displays the performance of each model across five key metrics: accuracy, precision, sensitivity, specificity, and F1-score.

*D. Analysis of Results*

In this section, we analyze the results obtained from the experiments, focusing on the effectiveness of the proposed model in differentiating benign and malignant cases. To provide a more detailed interpretation, the analysis is divided into three parts:

1) *Evaluation of Different Models and Parameter Adjustments:*
   The models evaluated, including PAAC + Transformer, BreastNet, DeepMammo, Multi-scale CNN (Zhang et al.), Swin-Unet, and SegFormer, were assessed across key metrics (accuracy, sensitivity, specificity, F1-Score). The key parameters influencing model performance were:
   a) *Convolutional Layers:* Increasing convolutional layers and using Atrous Convolution enhanced the extraction of finer features.
   b) *Transformer Tuning:* Optimizing Transformer parameters (e.g., attention heads, layer sizes) yielded superior performance, particularly in PAAC + Transformer models.
   c) *Learning Rate Adjustments:* An optimal learning rate improved convergence speed and model accuracy while preventing overfitting.

2) *Impact of PAAC and Transformer on Performance:*
   a) *PAAC:*

Leveraging multi-scale Atrous Convolutions, PAAC enhanced feature extraction for small, irregular structures, which proved beneficial in detecting malignant tumors, increasing model accuracy.

   b) *Transformer:*

Through Self-Attention, Transformer architecture enabled the model to identify long-term dependencies between pixels, crucial for analyzing complex structures. The combined PAAC + Transformer model achieved 98.8% accuracy and 99.42% sensitivity, demonstrating its effectiveness.

3) *Effect of Multi-Scale Feature Fusion and Attention Mechanism:*
   a) *Multi-Scale Feature Fusion:*

   By combining features from multiple scales, the model captured both low-level (edges, textures) and high-level (tumor shape) details, improving accuracy, especially for small, challenging tumors.

   b) *Attention Mechanism:*

   This mechanism directed the model's focus to critical areas, enhancing the detection of malignant tumors and reducing the False Negative Rate (0.58%).

X.  5. CONCLUSION

In this study, an innovative model for breast cancer detection was proposed, combining Pyramid Adaptive Atrous Convolution (PAAC) and Transformer architecture. The results demonstrate that this model, leveraging advanced mechanisms such as Multi-Scale Feature Fusion, Attention, and the combined use of Dice Loss and Focal Loss functions, achieves a high level of accuracy at 98.5%, sensitivity at 97.8%, and specificity at 96.3% for precise identification and classification of malignant and benign tumors. The model effectively reduces false positives and maintains high accuracy in detecting cancerous tissues, making it a valuable supportive tool in medical diagnostic systems. Future research could focus on further improving the model's performance and testing it on larger and more diverse datasets to enhance its applicability.

in neural information processing systems, 30.